\documentclass[lettersize,journal]{IEEEtran}
\usepackage{amsmath,amsfonts}
\usepackage{algorithmic}
\usepackage{algorithm}
\usepackage{array}
\usepackage[caption=false,font=normalsize,labelfont=sf,textfont=sf]{subfig}
\usepackage{textcomp}
\usepackage{stfloats}
\usepackage{url}
\usepackage{verbatim}
\usepackage{graphicx}
\usepackage{multirow} 
\usepackage{makecell}
\usepackage{amssymb}
\usepackage{booktabs}
\usepackage[numbers,sort&compress]{natbib}
\usepackage{hyperref}
\usepackage{soul}
\soulregister\ref{7}
\soulregister\cite{7}

\sethlcolor{lightyellow}

\hypersetup{hypertex=true,
            colorlinks=true,
            linkcolor=red,
            anchorcolor=green,
            citecolor=green}

\begin{document}
\definecolor{lightyellow}{RGB}{255, 249, 196}
\definecolor{qgblue}{RGB}{68,114, 196}
\definecolor{qgorange}{RGB}{237, 125, 49}
\definecolor{qgred}{RGB}{192, 0, 0}
\title{Bring Adaptive Binding Prototypes to Generalized Referring Expression Segmentation}

\author{Weize Li, Zhicheng Zhao, Haochen Bai, Fei Su
\thanks{This work was supported by the Chinese National Natural Science Foundation under Grant 62076033. \textit{(Corresponding author: Zhicheng Zhao.)}}
\thanks{The authors are with the Beijing Key Laboratory of Network System and Network Culture, Key Laboratory of Interactive Technology and Experience System Ministry of Culture and Tourism, School of Artificial Intelligence, Beijing University of Posts and Telecommunications, Beijing 100876, China (e-mail: bupt\_lwz@bupt.edu.cn; zhaozc@bupt.edu.cn; baihuplehpy@163.com; sufei@bupt.edu.cn)}}

\markboth{Journal of \LaTeX\ Class Files,~Vol.~14, No.~8, August~2021}%
{Shell \MakeLowercase{\textit{et al.}}: A Sample Article Using IEEEtran.cls for IEEE Journals}


\maketitle

\begin{abstract}
Referring Expression Segmentation (RES), which aims to identify and segment objects based on natural language expressions is garnering increased research attention. While substantial progress has been made in RES, the emergence of Generalized Referring Expression Segmentation (GRES) introduces new challenges by allowing the expressions to describe multiple objects or lack specific object references. Existing RES methods usually rely on sophisticated encoder-decoder and feature fusion modules, and have difficulty generating class prototypes that match each instance individually when confronted with the complex referent and binary labels of GRES. In this paper, reevaluating the differences between RES and GRES, we propose a novel Model with Adaptive Binding Prototypes (MABP) that adaptively binds queries to object features in the corresponding region. It enables different query vectors to match instances of different categories, or different parts of the same instance, significantly expanding the decoder's flexibility, dispersing global pressure across all the queries, and easing the demands on the encoder. The experimental results demonstrate that MABP significantly outperforms the state-of-the-art methods in all three splits on the gRefCOCO dataset. Moreover, MABP outperforms the state-of-the-art methods on the RefCOCO+ and G-Ref datasets, and achieves very competitive results on RefCOCO. The code is available at \href{https://github.com/buptLwz/MABP}{https://github.com/buptLwz/MABP}.
\end{abstract}
 

\section{Introduction}
Referring Expression Segmentation (RES) is one of the most challenging tasks in multimodal information processing. Given an image, and a natural language expression describing an instance in the image, RES aims to identify the corresponding object and generate a segmentation mask\cite{ding2021vision,hu2016segmentation}. RES has demonstrated significant application potential in various fields, such as human-robot interaction\cite{wang2019reinforced} and image editing\cite{chen2018language}. In recent years, substantial progress has been achieved, particularly on well-established datasets such as ReferIt\cite{kazemzadeh2014referitgame} and RefCOCO\cite{mao2016generation,yu2016modeling}. These studies adhere to the classical rules of RES, in which expressions only describe a unique instance. To further expand the application range of RES, an extension beyond the classical rules has led to the introduction of the multi-target RES dataset, gRefCOCO, and its corresponding benchmark, known as Generalized Referring Expression Segmentation (GRES)\cite{liu2023gres}. In contrast to RES, expressions within GRES may describe multiple objects or lack object references, presenting a new challenge to the RES.

\begin{figure}[t]
  \centering
  \includegraphics[width=1\linewidth]{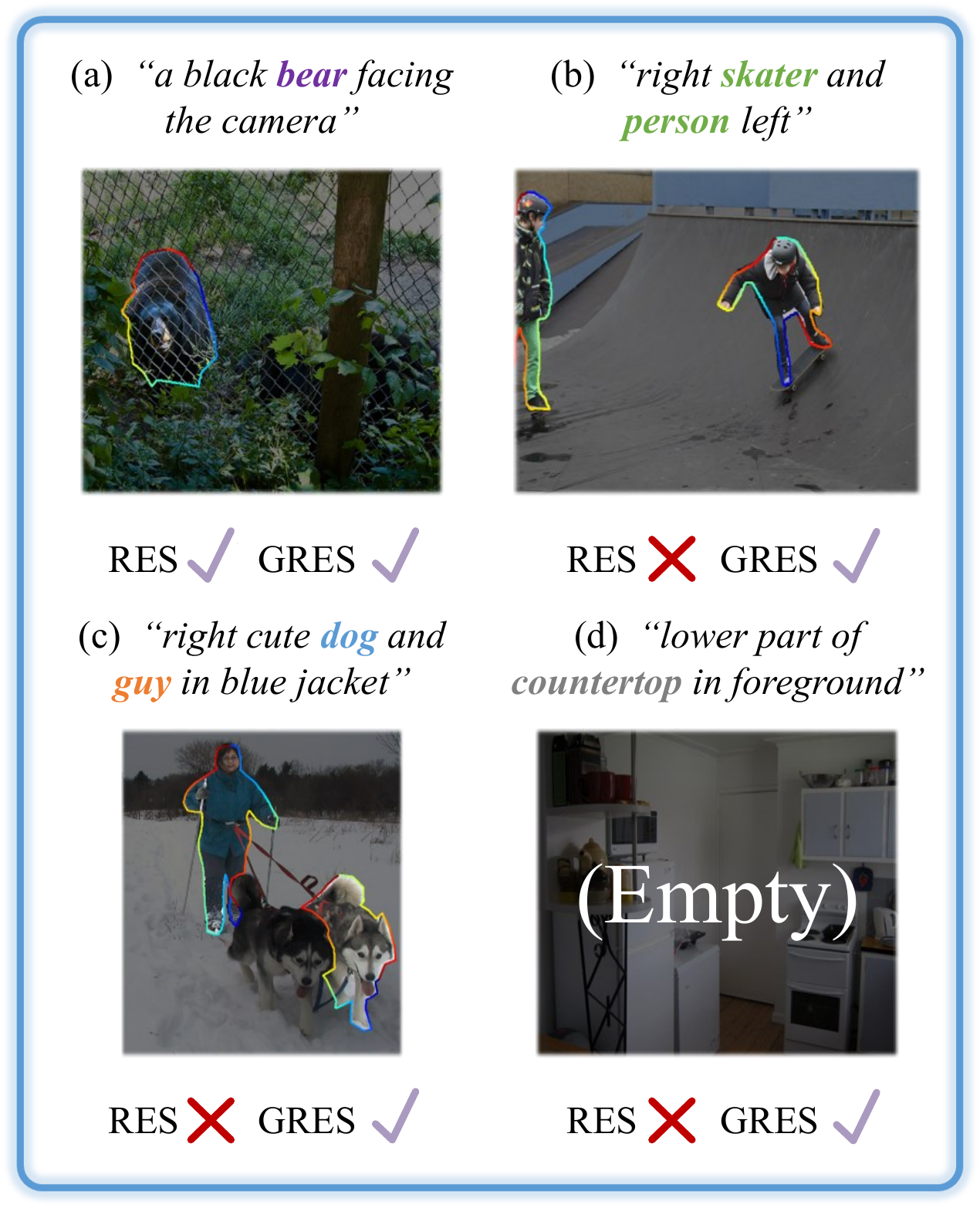}
  \centering
\caption{RES vs. GRES. The classic RES is designed to handle expressions that specify a single target object. In contrast, GRES extends this capability by supporting expressions that indicate an arbitrary number of target objects. For example, GRES accommodates multi-target expressions such as (b) and (c), as well as expressions indicating no target, as shown in (d). Notably, some multi-target expressions in GRES may even describe instances belonging to different classes, such as (c).}
   \label{fig:vs}
\end{figure}

Existing RES methods often use complex encoder-decoder systems and various feature fusion modules\cite{huang2020referring,yu2018mattnet} to build the classical segmentation paradigm. Recently, improvements in RES have been driven primarily by Transformers\cite{ding2021vision,9932025,kim2022restr,wang2022cris}, where a set of learnable query vectors are generated for each expression to serve as class prototypes for mask predictions.

These methods emphasize a greedy approach, aiming to generate unique feature prototypes for all the potential categories. However, these methods have achieved very limited success in GRES\cite{liu2023gres}. An intuitive issue is that, although the expression describes multiple targets, GRES only provides binary labels for the foreground and background, without distinguishing between different instances. The various combinations of instances greatly expand the potential number of classes in the dataset. In addition to different target quantities, as illustrated in Fig. \ref{fig:vs}, a deeper distinction between RES and GRES is that some instances in an expression are of the same category (Fig. \ref{fig:vs} (b)), whereas others belong to different categories (Fig. \ref{fig:vs} (c)). When faced with samples containing instances of multiple categories, RES methods are easily influenced by the prior knowledge of pre-trained models that distinguish features of instances from different categories, whereas the loss function expects the encoder to encode them into similar features. This contradiction obviously increases the learning burden on the model.

\begin{figure}[t]
  \centering
  \includegraphics[width=1\linewidth]{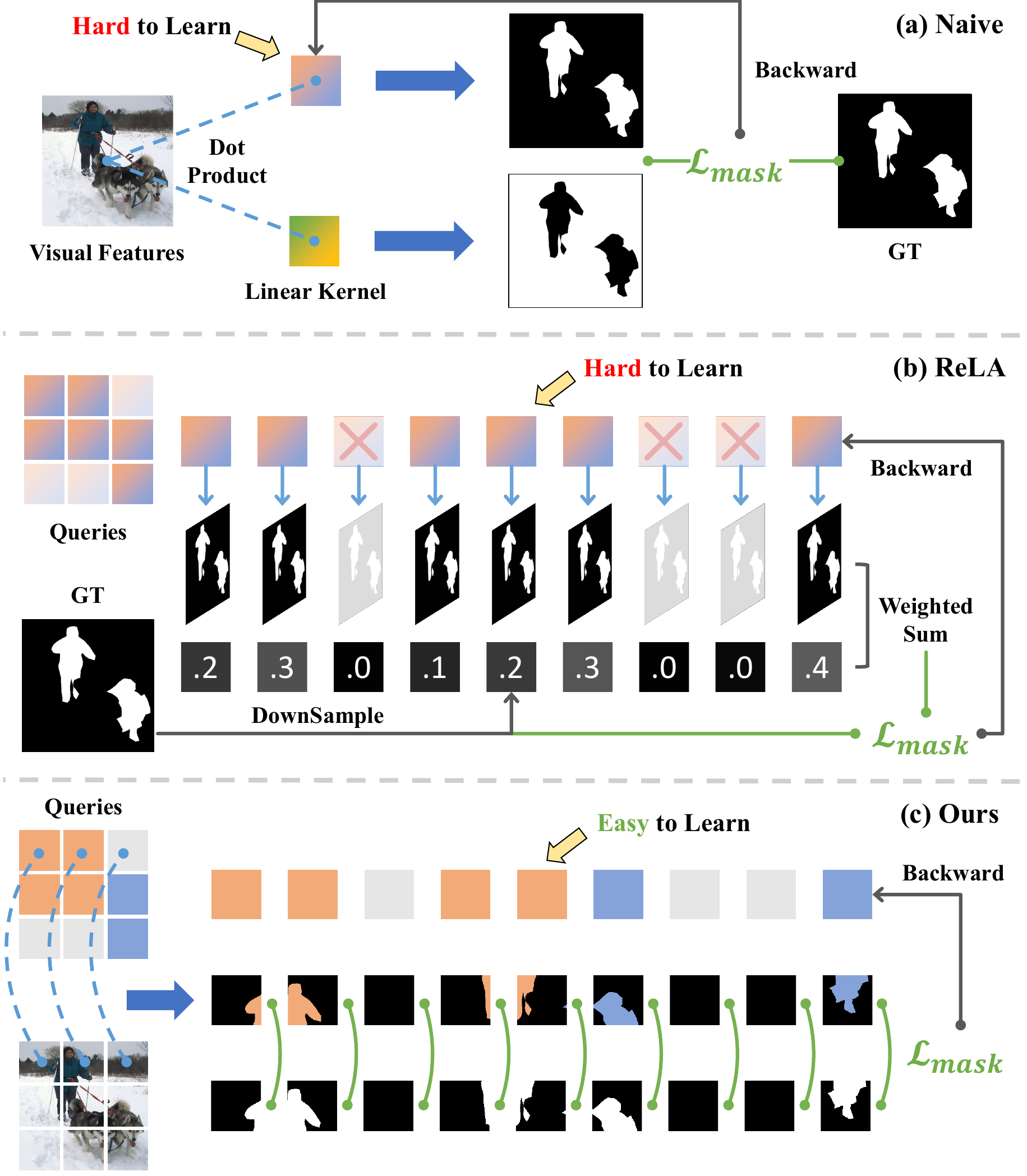}
  \centering
\caption{Comparison between the proposed adaptive binding prototypes and previous methods. (a) shows the naive pixelwise classification approach commonly used in segmentation, exemplified by a linear layer serving as the classification head. (b) ReLA's\cite{liu2023gres} mask head, uses downsampled ground truths (GTs) as weights to aggregate the prediction masks generated by multiple queries. (c) introduces the proposed adaptive binding prototype method. We divide the feature map into various regions and compute the loss separately, thereby constraining the queries to become more learnable class prototypes compared with the above two approaches.}
   \label{fig:kernel}
\end{figure}

As shown in Fig. \ref{fig:kernel}, taking a single sample as an example, most RES methods resemble the naive scenario described in Fig. \ref{fig:kernel} (a), which relies solely on a single class prototype (query in Transformer or convolution kernel) to summarize all the foreground targets. The ReLA\cite{liu2023gres} (Fig. \ref{fig:kernel} (b)) uses a downsampled version of the ground truth to control the proportion of each query's mask in the final output, allowing for a certain tolerance to the inherent bias in the pre-trained model. However, from a loss perspective, different queries in the ReLA only receive gradients of different scales but are still required to summarize all the targets. As a result, the multimodal features and queries all exhibit pronounced spatial consistency to ensure the recognition of every target. Designing distinct feature prototypes for different instances should address this issue well. However, applying these to GRES is nontrivial, as it only provides labels for the entire expression so that all references within the expression will be classified as foreground\cite{liu2023gres}. Therefore, it is challenging to guide queries to care about different objects without the annotation supervision of every individual instance mentioned in the referent.

To address these issues, as illustrated in Fig. \ref{fig:kernel} (c), instead of excavating queries corresponding to different targets, we divide the feature map into various regions, making queries adaptively bind to the target features of the corresponding region. We fully take advantage of the prior knowledge of the pre-trained models and facilitate the assignment of unique feature prototypes to different classes' instances or various regions within the same instance. From this prototype-based perspective, we propose a Model with Adaptive Binding Prototypes (MABP) for GRES, which consists of a query generator, a multimodal decoder (MMD), and a regional supervision head (RSH). Our tight binding with regions enables adaptive binding between different query vectors and instances of different categories, significantly expanding the decoder's flexibility, dispersing global pressure across all queries, and easing the heavy demands on the encoder.

In summary, our contributions are as follows:

\begin{itemize}
\item We propose a regional supervision head that effectively achieves adaptive alignment between prototypes and various class instances, which leads to an improved performance in complex task scenarios involving multiple class instances.
\item We introduce a mixed modal decoder that facilitates the interaction of multimodal features and context learning at a lower computational cost. This design has notable advantages, particularly in handling no-target samples in GRES.
\item We present a novel strategy of region-based queries via an end-to-end architecture that enables queries to bind with regions while maintaining the knowledge in the pre-trained model. Our approach outperforms the state-of-the-art (SOTA) models on three datasets on GRES and RES.

\end{itemize}

\section{Related Works}
\textbf{RES and GRES} aim to segment objects in images based on natural language references. Early works, such as \cite{hu2016segmentation}, initially followed classical segmentation paradigms by concatenating text features and visual features to obtain segmentation masks. The success of REC \cite{liao2020real,yang2020improving} inspired a series of two-stage methods \cite{9745353,yu2018mattnet}, where candidate boxes are extracted and text features are used to select the target instances. Recently, Transformer-based approaches \cite{9932025,ding2021vision,wang2022cris} have been proposed, and have made significant progress. For example, \cite{yang2022lavt,9932025} utilized the Swin Transformer as the visual encoder, aggregating text and visual features through attention modules and enhancing the localization capability. However, per-pixel classification was still followed until MaskFormer\cite{cheng2021per}, which led to the emergence of new mask classification-based methods \cite{tang2023contrastive,liu2023gres}. A mask is predicted for each potential instance in the image, and then a classification is performed at the mask level. Based on mask classification, CGFormer \cite{tang2023contrastive} stands out for incorporating contrastive learning.

However, the current RES focuses only on a ``one expression, one instance" scenario, limiting the extension of RES to more generalized real-world situations. Therefore, new datasets such as gRefCOCO\cite{liu2023gres} and group RES\cite{wu2023advancing} have been proposed, and trigger a new task, GRES. Research on GRES is still in the early stages but continues to draw inspiration from the achievements of classical RES. ReLA\cite{liu2023gres} uses a weight matrix to aggregate masks produced by a normal mask classification model and achieves competitive results on gRefCOCO. The weight matrix is supervised by downsampling the ground truth and selectively aggregating masks from the foreground region. Therefore, all queries from the foreground region are considered equivalent global class prototypes, which reduces flexibility. In addition, the success of large language models has brought new opportunities to RES and GRES. \cite{xia2024gsva} collected extensive datasets, and by pretraining and fine-tuning large language models, it achieved better results than the conventional approaches.

\textbf{Semantic and Instance segmentation.} The general semantic segmentation task can be described as the task of classifying each pixel in an image based on its visual semantics. FCN is considered one of the pioneering works \cite{long2015fully}, and it constructs a symmetric encoding-decoding network by stacking convolutional modules. Further developments include U-Nets\cite{ronneberger2015u,10.1007/978-3-030-00889-5_1} and Deeplabs\cite{Chen_2018_ECCV}, which aggregate multiscale feature maps, significantly improving the performance.

Unlike semantic segmentation, instance segmentation not only demands the distinguishing between various categories in an image but also discerning different instances within the same category. Its strong correlation with object detection has inspired a series of two-stage methods\cite{he2017mask,9745353}, which segment object instances from detection boxes and have multitasking capabilities. After that, inspired by deformable DETR\cite{zhu2020deformable} in object detection, MaskFormer\cite{cheng2021per} extended it to segmentation. Built upon mask classification, MaskFormer uses a Transformer decoder to facilitate interactions between a set of learnable queries and visual features, thereby predicting masks and classifying them. While Max-deeplab\cite{wang2021max} shares a remarkable similarity with MaskFormer, it applies Softmax and argmax to the output, ensuring no overlap between masks. While they have good performance, both require Hungarian matching to bind prototypes to the targets, and they discard queries with lower IoUs, thus suppressing model efficiency.

\textbf{Segmentation from the Prototype View.}
In contrast to mainstream segmentation strategies, \cite{zhou2022rethinking} proposed a segmentation framework based on a prototype view. Specifically, drawing inspiration from prototype learning, \cite{zhou2022rethinking} posits that the role of the encoder and decoder is to pull features of the same class closer, while pushing features of different classes farther. The segmentation head only measures the distance between features and prototypes to categorize each pixel. Similarly, \cite{10445499} shifted the focus of segmentation to prototypes, utilizing contrastive learning to update predefined class prototypes and accomplish segmentation. From a prototype view, the success of MaskFormer\cite{cheng2021per} can further substantiate the advancement of \cite{zhou2022rethinking,10445499}. The queries initialized in MaskFormer essentially serve as various feature prototypes, which engage in matching with the feature maps after decoding. Notably, MaskFormer allows overlapping output results, implying that the same instance can have multiple sets of prototypes, and align with \cite{zhou2022rethinking}. However, constrained by labels, \cite{zhou2022rethinking} and \cite{10445499} cannot effectively supervise the generation of prototypes; they can be predefined using clustering methods, resulting in fixed prototypes without adaptive capabilities.

\begin{figure*}[htbp]
    \centering
    \includegraphics[width=1\linewidth]{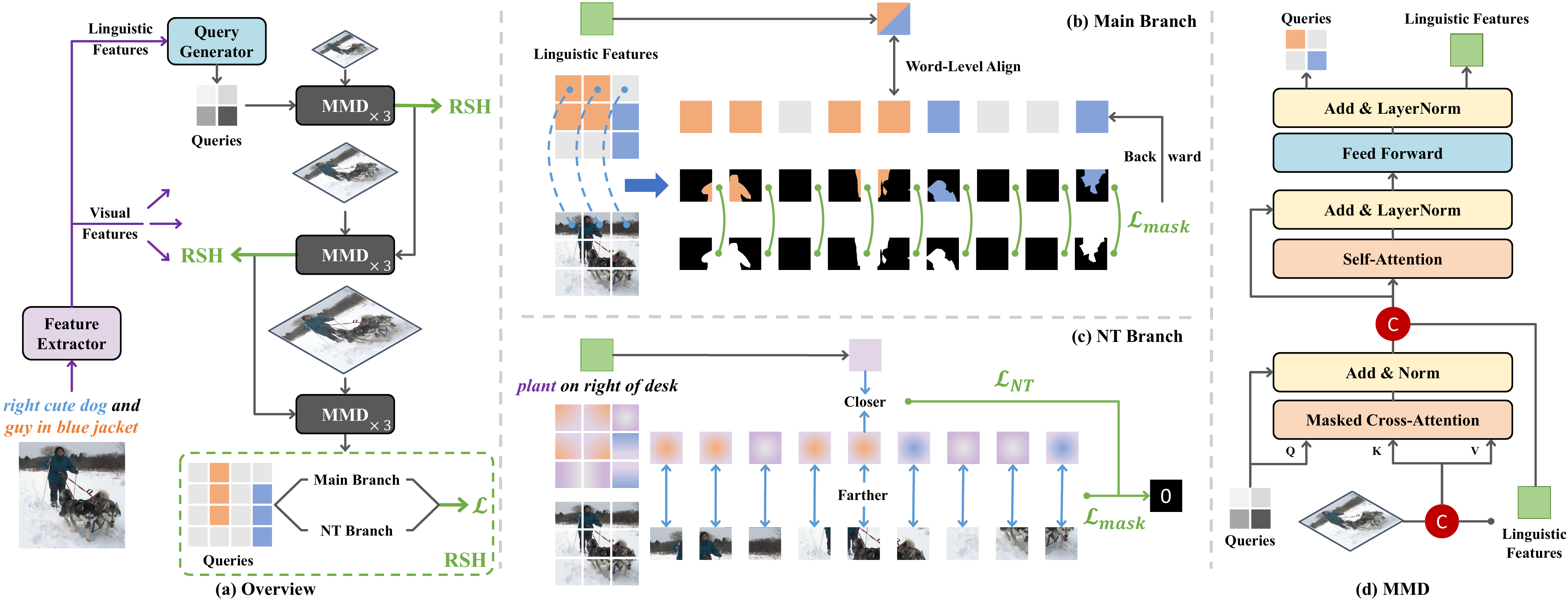}
\caption{The overall architecture of the proposed MABP. Initially, we utilize a feature extractor to obtain the linguistic features and visual features. The linguistic features are then combined with learnable region embeddings to generate region-text-specific queries via a query generator. Then, a set of mixed modal decoders (MMDs) are employed for these queries to interact gradually with visual features for reasoning. Finally, the decoded queries, along with visual and linguistic features, are fed into the Regional Supervision Head (RSH) to obtain prediction masks and no-target indicators. (b) and (c) illustrate the two branches of RSH, while (d) shows the detailed structure of the MMD.}
    \label{fig:archi}
\end{figure*}
\section{Method}

The entire framework of our proposed MABP is shown in Fig. \ref{fig:archi} (a). First, we adapt the feature extractor to encode both the images and reference expressions. The linguistic features are then fed into the query generator to combine with learnable embeddings, generating region-text-specific queries. The queries and linguistic features subsequently engage in multi-level interactions with visual features at various scales through the MMD. Each decoding layer consists of three sets of MMD and one RSH to obtain intermediate results at each scale for deep supervision. Unlike prior Transformer-based methods such as \cite{9932025,ding2021vision,tang2023contrastive,liu2023gres}, where queries typically maintain a fixed size and treat each feature map equally, we apply nearest upsampling to progressively increase the number of queries with improved of mask feature resolutions. Our framework emphasizes the relative stability of the feature prototypes during scale changes, while also increasing the flexibility of queries in learning the local details, and we allow queries to inherit knowledge learned at coarse granularities into the learning of fine-grained knowledge.

\subsection{Feature extractors}

\textbf{Visual Feature Extractor.} Following previous work\cite{yang2022lavt,tang2023contrastive,liu2023gres}, we employ the Swin Transformer \cite{liu2021swin} as the visual encoder. When given an input image $ I \in \mathbb{R}^{H\times W \times 3}$ with a size of $H \times W$, the encoder extracts its visual feature map at three stages, where each stage corresponds to an encoding block of the Swin Transformer with resolutions of 1/32, 1/16, and 1/8 of the original image. We then feed them into the pixel decoder for pixel-level decoding, which is constructed via the advanced multiscale deformable attention transformer decoder\cite{zhu2020deformable}.

\textbf{Language Encoder.} We employ BERT \cite{devlin2018bert} as the language encoder following \cite{yang2022lavt,tang2023contrastive,liu2023gres}. For an expression containing L words, we extract its linguistic feature, denoted as \(e \in \mathbb{R}^{C_L}\), where \(C_L\) is the channel dimension. Additionally, we acquire word representations by excluding the last pooling layer, represented as \(\mathbf{l} \in \mathbb{R}^{L \times C_{L}}\).

\subsection{Query generator}

In previous work \cite{9932025,ding2021vision,tang2023contrastive,liu2023gres}, influenced by DETR\cite{zhu2020deformable}'s achievements in object detection and semantic segmentation, the queries were often randomly initialized and repeated along the batch dimension. However, unlike in conventional tasks, the prior knowledge of categories carried by expressions means that GRES neither seeks nor can achieve greedy, universal feature prototypes, which makes the practice of sharing the initial queries in minibatches meaningless. Therefore, we construct a query generator to specialize in learnable embeddings for the different expressions and regions, thereby integrating prior information into the initial queries.

\begin{figure}[htbp]
    \centering
    \includegraphics[width=1\linewidth]{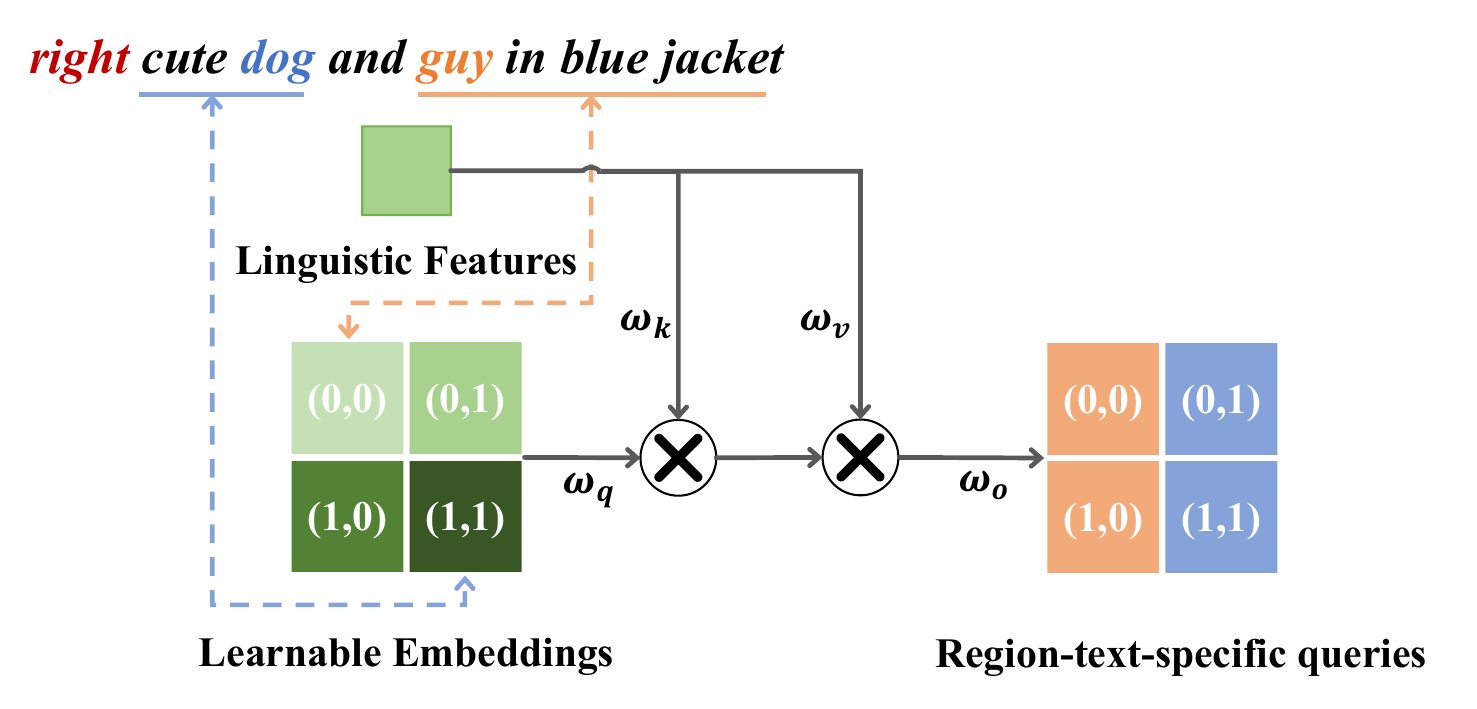}
\caption{The structure of the proposed query generator. Unlike traditional random initialization, our initialization query first undergoes cross-attention processing with linguistic features. For example, in the description ``{\color{qgred}{right}} cute {\color{qgblue}{dog}} and {\color{qgorange}{guy}} in blue jacket", the {\color{qgblue}{dog}} is obviously on the {\color{qgred}{right}} side of the {\color{qgorange}{guy}}, and the {\color{qgorange}{guy}} is on the left. Therefore, when our query generator is used, the query in the left region will integrate more information about the {\color{qgorange}{guy}}, whereas the query on the right will focus on information about the {\color{qgblue}{dog}}.}
    \label{fig:QG}
\end{figure}

As illustrated in Fig. \ref{fig:QG}, where $\omega_{q}$, $\omega_{k}$, $\omega_{v}$, and $\omega_{o}$ represent projection functions, the module first initializes a set of learnable embeddings $\mathbf{r} \in \mathbb{R}^{N_{0} \times C}$ for each region to capture prior position knowledge. Here, $N_{0}$ denotes the initial total number of regions. Given an expression such as ``right cute dog and guy in blue jacket'' and its features $\mathbf{l} \in \mathbb{R}^{L \times C_L}$ obtained from the language encoder, we use $\mathbf{r}$ as the query, and $\mathbf{l}$ as the key and value, to extract prior region-based information. Since the expression directly conveys spatial information that the ``dog'' is to the right of the ``guy'', the query in the right region will contain more information about the ``dog'', whereas the query in the left region will contain more information about the ``guy''. Therefore, by collecting word features of interest from various regions, the query generator finally forms region-text-specific initialization queries, $\mathbf{Q}_0$.

By pre-extracting category information within the expression, we significantly reduce the demands on learnable embeddings, allowing them to focus on globally shared position information and setting our model apart from typical DETR-based methods \cite{liu2023gres, zhu2020deformable}. Our subsequent modules then establish a strong binding between learnable embeddings and regions, enabling the embeddings to adaptively acquire cross-dataset knowledge for different regions.

\subsection{Mixed modal decoder}

Our MMD consists of three modules processed in the following order: a masked cross-attention module, a self-attention module, and a feed forward network (FFN). Owing to the no-target samples in GRES, when the query interacts solely with visual features, and there are no positive instances in the sample, the attention map must exhibit the most unstable uniform distribution state \cite{hyeon2023scratching}. Therefore, we extract the language features from the previous stage and incorporate them into the decoding stage. As illustrated in Fig. \ref{fig:archi} (d), it concatenates with the visual features in the masked cross-attention module and with the query in the self-attention module. Serving not only as positive instance placeholders but also facilitating context learning in self-attention. Specifically, given the visual features $\mathbf{V}$, queries $\mathbf{Q}_{i}$, attention mask $\mathbf{M}$ and linguistic feature $\mathbf{l}_i$, we obtain $\mathbf{Q}_{i+1}$ via the following equation:
\begin{equation}
    \begin{aligned}
        \mathbf{S}_i &= \mathbf{V}\oplus \mathbf{l}_i \notag \\
        \mathbf{Q}_{i+1} &= \text{Softmax}(\mathcal{M}+ \mathbf{Q}_i\mathbf{S}_i^T)\mathbf{S}_i + \mathbf{Q}_i \\
    \end{aligned}
\end{equation}
\begin{equation}
    \begin{aligned}
        \mathbf{X}_i &= \mathbf{Q}_{i+1}\oplus \mathbf{l}_i \\
        \mathbf{X}_{i+1} &= \text{Softmax}(\mathbf{X}_i\mathbf{X}_i^T)\mathbf{X}_i + \mathbf{X}_i \\
        (\mathbf{Q}_{i+1},\mathbf{l}_{i+1}) &= \text{Split}(\text{FFN}(\mathbf{X}_{i+1}))
    \end{aligned}
\end{equation}
where $\oplus$ means concatenating along the first dimension. The ``Split" implies slicing the fusion features $X_{i+1}$ along the first dimension based on the shapes of $\mathbf{Q}_{i+1}$ and $\mathbf{l}_{i+1}$. The attention mask $\mathcal{M} \in \{-\infty,0\}$ is used to control the receptive field of the cross-attention module, enabling queries to ignore unnecessary regions and improving computational efficiency.

\subsection{Regional supervision head}

\subsubsection{Main Branch} 

\label{sec:MB}

As discussed above, we aim to extract the feature prototypes corresponding to the instance categories in the regions where the instances are located. The carriers of these prototypes are the queries of the corresponding regions. In DETR\cite{zhu2020deformable}-based methods, queries are typically specialized into class prototypes through Hungarian matching, which is achieved by separately propagating the gradients generated from different categories back to their corresponding queries, which is not feasible in GRES with only binary labels. To achieve a similar effect in GRES with only implicit instance differentiation, we propose dividing the visual features and their corresponding ground truths into multiple patches based on region size, ensuring that each patch contains instances of only a single category as much as possible, which is simple but efficient. We establish strong bindings between queries and each patch so that gradients generated by each patch can only propagate to the corresponding queries, which specializes queries into prototypes of a single category. Furthermore, the proposed query generator and MMD extract prior category information from the linguistic features, assisting RSH in adaptively binding queries to the class feature prototypes mentioned in the expression.

Specifically, as shown in Fig. \ref{fig:archi} (b), we partition the feature map $\mathbf{V}$ of the current layer into $N_{i}$ patches via a simple sliding window operation, where $N_{i}$ represents the total number of regions in the current layer. Next, we perform matrix multiplication separately between each query and its corresponding patch to obtain a set of binary prediction masks. The resulting masks have a shape of $H_{win} \times W_{win} \times N_i \times C$. Here, $H_{win}$ and $W_{win}$ denote the height and width of each sliding window, respectively, and they can be computed as $(H_{win},W_{win}) = \lfloor (H,W) / \sqrt {N_{i}} \rfloor +1$. After applying sigmoid activation, the output region mask is subsequently used to compute the loss \(L_{\text{mask}} \) with the ground truth.

\subsubsection{No-target (NT) Branch} Designing a separate No-Target Indicator for no-target samples is crucial. For example, in \cite{liu2023gres}, an MLP with two hidden layers was used to directly map the queries to the indicator. However, relying solely on queries is clearly insufficient to determine the presence or absence of targets, especially when the MMD to provide linguistic information for the queries is lacking. Unlike \cite{liu2023gres}, we construct a novel triplet-based approach that involves queries, linguistic features, and visual features. We consider that, as the carriers of class prototypes, to avoid matching positive instances in the feature map, queries should evidently be closer to the linguistic features than to the visual features.

Therefore, as shown in Fig. \ref{fig:archi} (c), given the visual feature map $\mathbf{V}$ at the current scale, the NT branch first applies average pooling to obtain feature centroids for each region, with a shape of $N_i \times C$. Then, for linguistic features $\mathbf{l}_i$, we use an MLP to aggregate word-level features into sentence-level linguistic embeddings, which are transformed to the same size as the queries. The queries are dot-producted separately with the pooled visual features and linguistic embeddings to obtain two sets of similarity matrices. After these two similarity matrices are concatenated, a final no-target indicator is obtained through an MLP with two hidden layers, and is utilized to calculate \(L_{\text{NT}} \). Notably, in this process, we still apply the same \(L_{\text{mask}} \) to no-target samples, where the ground truth is an all-zero sample.

\subsection{Loss Compute}
For \(L_{\text{mask}} \), we simultaneously calculate its cross-entropy loss and Dice loss \cite{li2019dice} with the ground truth $\mathbf{Y}$ via
\begin{equation}
\begin{split}
    \mathbf{L}_{CE}^i = -\mathbf{Y}_i \cdot \log [\sigma(\mathbf{O}_i)]-(1-\mathbf{Y}_i)\cdot \log[1-\sigma(\mathbf{O}_i)]
\end{split}
\end{equation}

\begin{equation}
    \mathbf{L}_{DE}^i = 1-[\frac{2 \mathbf{Y}_i\cdot \sigma(\mathbf{O}_i)+\epsilon}{\mathbf{Y}_i+\sigma(\mathbf{O}_i)+\epsilon}]
\end{equation}
where $\sigma$ represents the activation function, typically a sigmoid function, and where $\epsilon$ is a smoothing factor. $\mathbf{Y}_i$ is obtained from the ground truth $\mathbf{Y}$ through the nearest downsampling, ensuring that its shape is consistent with the output $\mathbf{O}_i$.

The final \(L_{\text{mask}} \) can be represented as:
\begin{equation}
\begin{aligned}
        L_{mask} =&  \sum_i (\omega_{ce} \mathbf{L}_{CE}^i+\omega_{de} \mathbf{L}_{DE}^i), \\
\end{aligned}
\end{equation}
where $\omega_{ce}$ and $\omega_{de}$ are used to adjust the proportions of the cross-entropy loss and Dice loss, respectively. For $L_{NT}$, we only applied cross-entropy loss for optimization.

\section{Results}

In this section, we conduct an experimental evaluation and performance comparison of the MABP. We first introduce the datasets, experimental details, and metrics, and compare MABP with seven methods. Then, we validate the effectiveness of the different strategies through ablation experiments.

\begin{table*}[htbp]
    \centering
    \caption{Results on classic RES in terms of cIoU.}

    \label{tab:res}
    \begin{tabular}{c|l|c|c|ccc|ccc|ccc} \toprule  
         \multirow{2}{*}{} & \multirow{2}{*}{Methods}  &\multirow{2}{*}{\makecell{Visual\\Encoder}}&\multirow{2}{*}{\makecell{Textual\\Encoder}}&  \multicolumn{3}{c|}{RefCOCO}&\multicolumn{3}{c|}{RefCOCO+} & \multicolumn{3}{c}{G-Ref}\\  
           &   &&& val&testA& testB&val& testA&testB & val-U& test-U&val-G\\ \midrule \midrule
 \multirow{10}{*}{\makecell{RESMethods}}& MCN~\cite{luo2020multi} &Darknet53& bi-GRU& 62.44& 64.20& 59.71& 50.62& 54.99&44.69 & 49.22& 49.40&-\\ 
         &  VLT~\cite{9932025}&Darknet53&bi-GRU&  65.65&  68.29& 62.73&55.50&  59.20&  49.36 & 52.99& 56.65&49.76\\   
         &  ReSTR~\cite{kim2022restr} &ViT-B&Transformer&  67.22&  69.30& 64.45&55.78&  60.44&  48.27 & -& -&54.48\\   
         &  CRIS~\cite{wang2022cris}&CLIP-R 101&CLIP&  70.47&  73.18& 66.10&62.27&  68.08&  53.68 & 59.87& 60.36&-\\   
         &  LAVT~\cite{yang2022lavt}&Swin-B&BERT&  72.73&  75.82& 68.79&62.14&  68.38&  55.10 & 61.24& 62.09&60.50\\
 & CM-MaskSD~\cite{10413654}& CLIP-ViT B& CLIP& 72.18& 75.21& 67.91& 64.47& 69.29& 56.55& 62.67& 62.69&-\\
 & VLT~\cite{ding2021vision}&Swin-B& BERT& 72.96& 75.96& 69.60& 63.53& 68.43&56.92 & 63.49& 66.22&62.80\\
 & CrossVLT~\cite{10345690}& Swin-B& BERT& 73.44& 76.16& 70.15& 63.60& 69.10& 55.23& 62.38& 63.75&-\\
 & BKINet~\cite{10227590}& CLIP-R 101& CLIP& 73.22& 76.43& 69.42& 64.91& 69.88& 53.39& 64.21& 63.77&61.64\\  
 & CGFormer~\cite{tang2023contrastive}&Swin-B&BERT& \textbf{74.75}& \textbf{77.30}& 70.64&64.54&71.00&57.14 & 64.68& 65.09&62.51\\ \midrule
         \multirow{2}{*}{\makecell{GRES\\Methods}}&  ReLA\textsuperscript{\dag}~\cite{liu2023gres} &Swin-B&BERT&  73.47&  76.60& 70.04&64.41&  69.18&  54.97 & 65.00& 65.97&62.70\\ 
         &  Our MABP&Swin-B&BERT&  \underline{74.48}&  \underline{76.73}  & \textbf{71.07}&\textbf{65.99}& \textbf{71.76}  & \textbf{57.22}   &\textbf{65.38} &\textbf{66.64} &\textbf{62.84}\\ \bottomrule

    \end{tabular}
\end{table*}

\begin{table}[htbp]
    \centering
    \caption{Comparison on gRefCOCO dataset.}

    \label{tab:gres}
    \begin{tabular}{l|cc|cc|cc} \toprule
          \multirow{2}{*}{Methods}&  \multicolumn{2}{c|}{Val}&  \multicolumn{2}{c|}{TestA}&  \multicolumn{2}{c}{TestB}\\  
          & cIoU&gIoU& cIoU&gIoU& cIoU&gIoU\\ \midrule  \midrule  
           MattNet~\cite{yu2018mattnet}&  47.51&  48.24&  58.66&  59.30&  45.33& 46.14\\ 
           LTS~\cite{jing2021locate}&  52.30&  52.70&  61.87&  62.64&  49.96& 50.42\\   
            VLT~\cite{9932025}&  52.51&  52.00&  62.19&  63.20&  50.52& 50.88\\   
           CRIS~\cite{wang2022cris}&  55.34&  56.27&  63.82&  63.42&  51.04& 51.79\\   
          LAVT~\cite{yang2022lavt}&  57.64&  58.40&  65.32&  65.90&  55.04& 55.83\\  
  CGFormer~\cite{tang2023contrastive}& 62.28 & 63.01 &68.15 &70.13 &60.18 &61.09\\ \midrule  
           ReLA~\cite{liu2023gres}&  64.20&  65.50&  70.78&  70.89&  60.97& 61.05\\ 
           MABP&  \textbf{65.69}&  \textbf{68.79}&  \textbf{71.60}&  \textbf{72.79}&  \textbf{62.75}& \textbf{64.01}\\ 

           \bottomrule

    \end{tabular}
\end{table}

\subsection{Datasets and Implementation Details}

\textbf{Experimental details} We conduct experiments on the gRefCOCO \cite{liu2023gres} dataset and classic RES datasets, including RefCOCO \cite{yu2016modeling}, RefCOCO+ \cite{yu2016modeling}, and G-Ref \cite{nagaraja2016modeling,mao2016generation}. All these datasets are based on MSCOCO \cite{lin2014microsoft} but are annotated according to different rules. gRefCOCO comprises 278,232 expressions, including 80,022 multiobject and 32,202 no-target samples. It utilizes 19,994 images, containing 60,287 unique instances. Other single-object datasets, RefCOCO and RefCOCO+, are smaller in scale, with only 120K references. The average text length in RefCOCO is 3.5 words, whereas in G-Ref, it is 8.4 words. RefCOCO+ restricts the use of absolute positional information for the reference targets. The datasets are split into training, validation, testA, and testB sets, following previous work.

\textbf{Implementation details}. Following \cite{liu2023gres}, our visual encoder is pretrained on ImageNet22K \cite{deng2009imagenet}, and the text encoder is initialized with HuggingFace weights \cite{wolf2020transformers}. The images are resized to 480x480. We employ AdamW \cite{loshchilov2017decoupled} with an initial learning rate of 2e-5 as the optimizer and train for 30 epochs with a batch size of 42. All the experiments are conducted on 6 NVIDIA A5000 GPUs. The evaluation metrics include the gIoU, cIoU, and precision at IoU thresholds of 0.7, 0.8, and 0.9, respectively, following \cite{liu2023gres}. The initial number of regions is set to \(N_0 = 16 \). Each time the visual feature scale doubles, the number of regions is quadrupled to ensure consistency.

\subsection{Comparison with State-of-the-Art Methods}

\textbf{Comparison on GRES} In Table \ref{tab:gres} and Table \ref{tab:res}, we present a comparative analysis of MABP against the SOTA methods on the GRES, as well as a comparison with the SOTA methods on classic RES. We reimplement the CGFormer and train it on gRefCOCO. To enhance no-target identification, output masks with fewer than 50 positive pixels are reset to all-negative.

Our MABP outperforms the SOTA methods on all splits of gRefCOCO, showing substantial improvement over the single-object models. Compared with ReLA\cite{liu2023gres}, MABP also achieves significant improvements, with a margin of 2\% for cIoU and 3\% for gIoU across the three splits of gRefCOCO. This finding indicates that MABP can effectively adapt to scenarios with multiple instances in the GRES, demonstrating the effectiveness of our proposed adaptive binding strategy. Furthermore, we evaluate the performance on the no-target samples. As shown in Table \ref{tab:nt}, our MABP outperforms the SOTA models by 7.04\% on N-acc, and achieves the second-best result on T-acc. Owing to the significantly larger number of positive samples than negative samples, T-acc has only a slight impact on the final results, and approaches its upper limit under non-overfitting conditions.

\begin{table}[htbp]
\centering
                
    \caption{No-target results comparison on GRES.}
    \label{tab:nt}
    \begin{tabular}{c|c|c|c} \toprule 
   &Methods& N-acc.& T-acc.\\ \midrule \midrule 
\multirow{4}{*}{\makecell{RES\\Methods}}&MattNet~\cite{yu2018mattnet}&  41.15 &96.13\\ 
            &VLT~\cite{9932025}&  47.17 &95.72\\  
            &LAVT~\cite{yang2022lavt}&   49.32&96.18\\ 
            &CGFormer~\cite{tang2023contrastive}&   51.01&96.23\\ \midrule 
\multirow{2}{*}{\makecell{GRES\\Methods}}&ReLA~\cite{liu2023gres}&   57.51&\textbf{96.97}\\  
   &Our MABP& \textbf{64.55}& \underline{96.40}\\ \bottomrule
    \end{tabular}
\end{table}

Table \ref{tab:pr} provides a comparison of MABP with ReLA and CGFormer on the Pr@0.9, 0.8, and 0.7 metrics, showing that MABP outperforms ReLA by 2.61\%, 1.91\%, and 1.3\%, respectively. The most notable improvement is observed for Pr@0.9, indicating that MABP excels in segmenting target details and small objects. This finding validates that our region-based framework can better perceive local information while retaining the global context, highlighting its heightened flexibility.

\begin{table}[htbp]
    \centering
    \caption{Pr results comparison on GRES.}
    \label{tab:pr}
    \begin{tabular}{c|c|c|c|c|c}\toprule 
         Methods&   Pr@0.9
&  Pr@0.8&  Pr@0.7
&  cIoU& gIoU\\ \midrule 
         CGFormer~\cite{tang2023contrastive}&   22.43
&  56.57&  68.93
&  62.28& 63.01\\
         ReLA~\cite{liu2023gres}&   23.56
&  57.01&  69.15
&  64.20& 65.50\\
         Our MABP&   \textbf{26.17}& \textbf{58.92}& \textbf{70.45}&  \textbf{65.69}& \textbf{68.79}\\ 
         \bottomrule
    \end{tabular}
\end{table}

\textbf{Comparison on Classic RES} To assess the generalization capabilities of MABP in handling single-object tasks, we show a comparison with the SOTA methods on classic RES in Table \ref{tab:res}. \textsuperscript{\dag} indicates that the results are reproduced in our environment according to the original configuration. Our MABP outperforms the SOTA methods on RefCOCOp and G-Ref, and achieves performances close to that of the SOTA methods on RefCOCO. This finding suggests that our MABP has a significant advantage in dealing with complex expressions, and exhibits superior generalizability on single-object datasets. This finding also indicates that, in addition to employing different prototypes for multiple object categories, adaptively using distinct prototypes for different parts of the same object contributes to improving the segmentation performance.

\subsection{Ablation Study}
\begin{figure*}[htbp]
    \centering
    \includegraphics[width=1\linewidth]{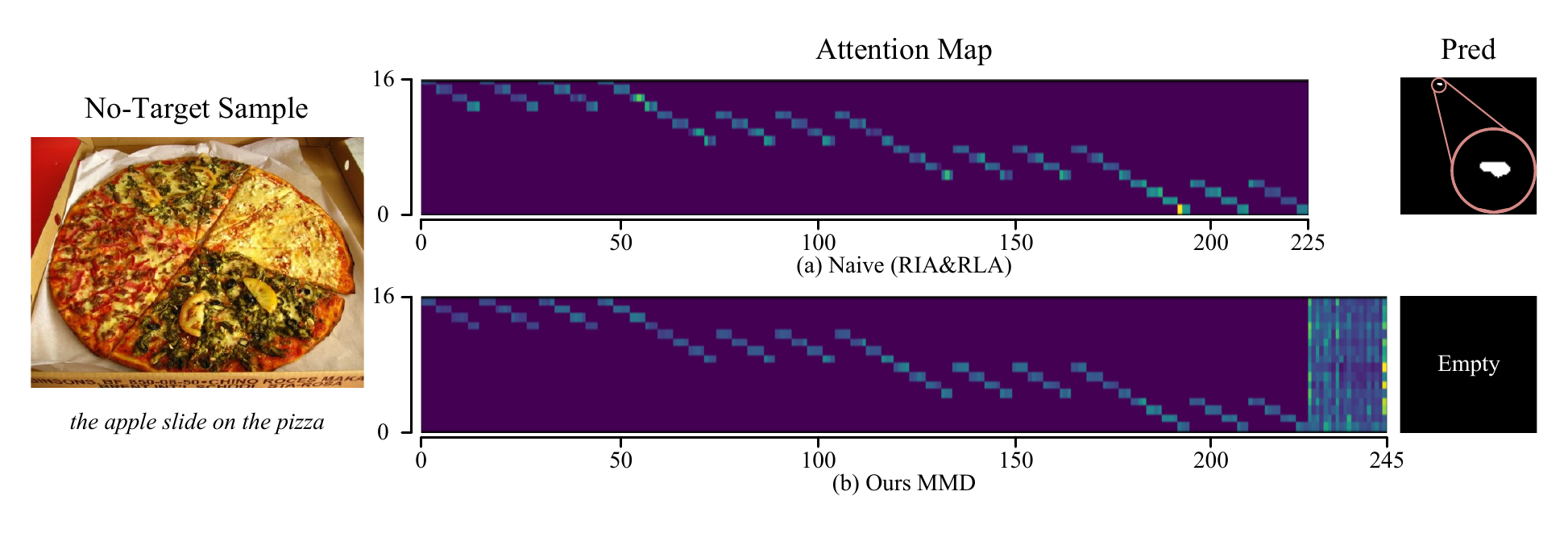}
\caption{Visualizations of the attention maps for the third-layer cross-attention module in the decoder. We input the same no-target sample into both the No. 2 model in Table \ref{tab:ablation} and our model, visualizing the cross-attention matrices of the decoder's third layer, i.e., the decoding module before the first mask head. As our mixed modal decoder incorporates linguistic features as placeholders, our model can learn a more easily interpretable non-uniform attention map, achieving better recognition for no-target samples.}
    
    \label{fig:attn}
\end{figure*}

\begin{table}[htbp]
    \centering
    \caption{Ablation study of MABP.}
    \label{tab:ablation}
    \begin{tabular}{c|l|c|c|c|c|c}\toprule
          No.&Methods  &cIoU &gIoU
&  Pr@0.9 
&  Pr@0.8&Pr@0.7
\\ \midrule
          1&Naive &63.07&64.23&  23.22& 56.89&69.01\\
          2&1+RSH&64.87&67.06& 24.87& 57.92&69.89\\
          
  3&2+MMD&65.21&67.94& 25.98& 58.15&70.13\\
  4&3+QG (full)&65.69&68.79& 26.17 & 58.92&70.45\\ \bottomrule
    \end{tabular}
\end{table}

In Table \ref{tab:ablation}, for a better comparison, we construct a naive model in No. 1, where only the main components of the architecture are retained, including randomly initialized queries, a Transformer decoder, and a naive mask head. We use the RIA and RLA modules from ReLA\cite{liu2023gres} as the Transformer decoder, and a naive mask head simply aggregates the prediction masks generated by the \(N \) queries into a final result. For No. 2 in Table \ref{tab:ablation}, we replace the naive mask head with the proposed RSH, achieving a significant improvement of 1.8\% in the cIoU and 2.83\% in the gIoU, demonstrating the effectiveness of the adaptive binding prototypes. Compared with Pr@0.7 and Pr@0.8, the improvement at Pr@0.9 is more notable, highlighting the advantage of RSH in extracting small targets and fine details.

For No. 3, we replace RIA and RLA with the proposed MMD, achieving further improvements in the cIoU and gIoU. For clarity, as depicted in Fig. \ref{fig:attn}, we visualize the masked cross-attention maps before the first output head for both the model in No. 2 of Table \ref{tab:ablation} and our MABP when facing a no-target sample. The result at ``Pred" has not undergone judgment from the no-target branch. Since there are no positive instances in the no-target samples, queries should be distanced from all the features to predict an empty result. In Fig. \ref{fig:attn} (a), which uses "RIA\&RLA" as the Transformer decoder, the attention map tends to have a uniform distribution. However, such a distribution is unstable for attention matrices \cite{hyeon2023scratching}, leading to inevitably high values in some regions, and resulting in incorrect patches in the output. In contrast, in our MMD (Fig. \ref{fig:attn} (b)), as linguistic features are involved in attention computations, even though there are no positive instances in the visual features, queries can treat linguistic features as positive for learning. This ensures a more manageable non-uniform distribution, maintaining an empty predicted result.

Finally, for No. 4 of Table \ref{tab:ablation}, we incorporate the query generator after the random initial queries to increase their flexibility. Compared with random initializations, our method achieves a better performance, indicating that our region-text-specific queries are a more favorable choice than the randomly initialized queries.

\begin{table}[htbp]

    \centering
    \caption{Ablation study of supervision depth.}
    \label{tab:supervision}
    \begin{tabular}{c|c|c|c|c|c}\toprule
          Branch& {\makecell{Supervision\\Depth $X$}}&cIoU &gIoU&  N-acc. & T-acc.
\\ \midrule
          Main& 1 &64.73&67.67&  62.21& 96.42\\
          \midrule
          \multirow{2}{*}{\makecell{NT}}&0&64.08&66.49& 58.26& 97.64\\
          
  &1&64.90&67.36& 60.40& 97.03\\
  \midrule
  Our MABP&3&65.69&68.79& 64.55 & 96.40\\ \bottomrule
    \end{tabular}
\end{table}

As mentioned earlier, influenced by classic networks such as \cite{10.1007/978-3-030-00889-5_1,Chen_2018_ECCV}, we adopt a deep supervision training strategy for MABP. To further verify the effectiveness of deep supervision, we conduct an ablation study on the number of supervision layers in Table. \ref{tab:supervision}. Here, a supervision depth of $X$ means that $X$ heads are used to supervise the outputs of the last $X$ layers. $X=0$ means that the branch is removed, and instead, we classify samples with fewer than 50 positive pixels in the output mask as NT. ``Main Branch" indicates that the ablation is applied only to the main branch, whereas the NT branch remains unchanged, and the ``NT Branch" is the opposite.

For the main branch, deep supervision shortens the supervision path, alleviating the vanishing gradient in the deep networks, which shows an improvement in the IoU compared with supervising only the last layer. For the NT branch, supervising only the final layer results in a noticeable drop in all the metrics except for T-acc. When the NT branch is entirely removed ($X=0$), the performance is further degraded. This is attributed to the long-tail distribution of the NT samples in the dataset, which leads to a strong bias in the main branch toward the target-present samples, and the gradient vanishing caused by the single NT branch further exacerbates this problem, weakening the network's ability to handle the NT samples effectively.

\subsection{Results of Video-based Referring Segmentation}

To further explore the upper bound of MABP, we perform additional experiments focused on refer video segmentation (RVOS). Although RVOS shares a similar setup with RES, its expressions emphasize motion information, thus this presents a considerable challenge to image-based models, including our MABP, which are originally limited to using single-frame data from videos.

We make some necessary modifications to the standard RVOS setup to match the MABP. Specifically, each frame in the videos is treated as an independent image, and during each iteration, only one frame is sampled from five adjacent frames to reduce training computational overhead. Regarding the dataset, we use MeViS\cite{ding2023mevis}, which strengthens the description of the motion states. Meanwhile, three metrics ($\mathit{J}$, $\mathit{F}$, $\mathit{J}\&\mathit{F}$) are applied to evaluate performance in accordance with \cite{ding2023mevis}.

\begin{table}[htbp]
\centering
    \caption{The results of MABP on MeViS}
    \label{tab:mevis}
    \begin{tabular}{c|c|ccc}
    \toprule
        Dataset & Methods & $\mathit{J}\&\mathit{F}$ & $\mathit{J}$ & $\mathit{F}$ \\ \midrule
        \multirow{9}{*}{\makecell{MeViS\\Val}} & URVOS~\cite{seo2020urvos} & 27.8 & 25.7 & 29.9 \\ 
        ~ & LBDT~\cite{ding2022language} & 29.3 & 27.8 & 30.8 \\
        ~ & MTTR~\cite{botach2022end} & 30.0 & 28.8 & 31.2 \\ 
        ~ & ReferFormer~\cite{wu2022language} & 31.0 & 29.8 & 32.2 \\
        ~ & VLT+TC~\cite{9932025} & 35.5 & 33.6 & 37.3 \\ 
        ~ & LMPM~\cite{ding2023mevis} & 37.2 & 34.2 & 40.2 \\ 
        ~ & HTR~\cite{10572009}& 42.7 & 39.9 & 45.5 \\ 
        ~ & DsHmp~\cite{he2024decoupling} & \textbf{46.4} &\textbf{43.0} &\textbf{49.8} \\ 
        ~ & Our MABP  & \underline{43.4} & \underline{39.9} & \underline{46.9} \\ \midrule
        \multirow{2}{*}{\makecell{MeViS\\Val-u}} & LMPM~\cite{ding2023mevis}& 40.2 & 36.5 & 43.9 \\ 
        ~ & Our MABP & \textbf{50.1} &\textbf{46.2} & \textbf{54.7} \\ \bottomrule
    \end{tabular}
\end{table}

The results are shown in Table \ref{tab:mevis}. Owing to its advantages in handling complex text, MABP still outperforms most video-based methods and achieves suboptimal results. However, it falls behind the SOTA DsHmp\cite{he2024decoupling} by 3\%. The results indicate two aspects. On the one hand, MABP can handle many scenarios in RVOS, demonstrating its generalizability. But on the other hand, MABP is not specifically proposed for video task. Therefore, in order to achieve better performance, it is necessary to introduce new measures.
 
\subsection{Visualization}
\label{sec:vis}
\begin{figure*}[htbp]
    \centering
    \includegraphics[width=1\linewidth]{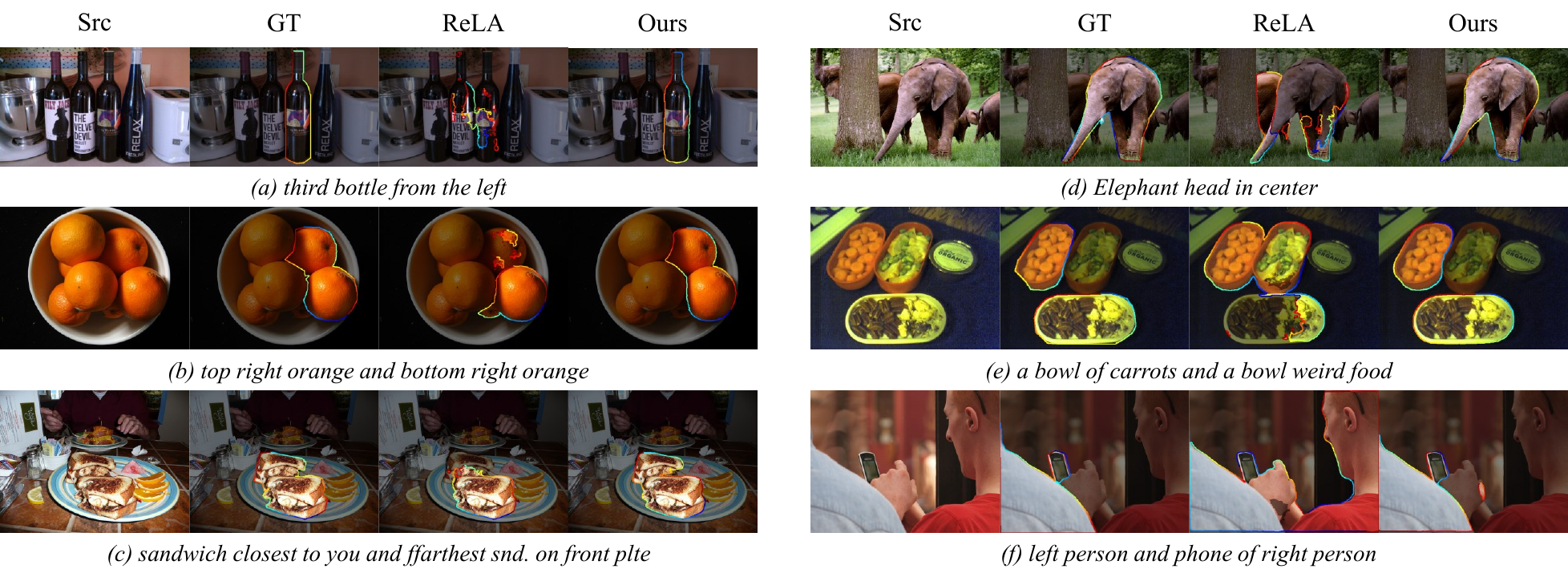}
\caption{Example results of our method on the gRefCOCO dataset. (a) and (d) represent single-target samples, whereas the others are multi-target samples. (f) illustrates a more challenging sample}
    \label{fig:vis}
\end{figure*}

\begin{figure}[htbp]
    \centering
    \includegraphics[width=0.9\linewidth]{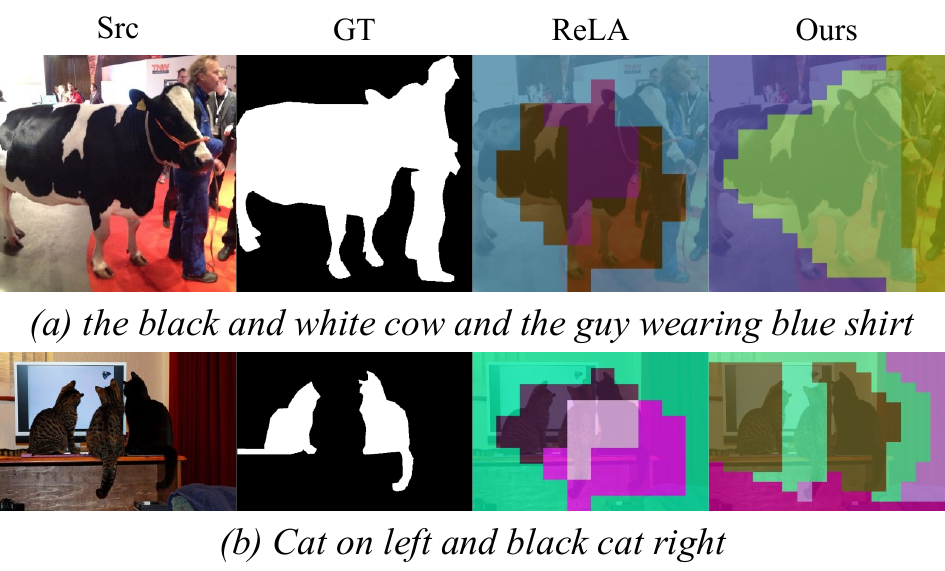}
\caption{Visualization comparison of query distributions. First, we separately extract the mask embeddings used by ReLA and our MABP in the final mask head. We subsequently use the K-means algorithm to cluster the embeddings. In the resulting images, blocks of the same color indicate that the embeddings from these regions are clustered into the same category. Our method is better than ReLA. Note, that since both ReLA and our MABP are region-based, these embeddings have a one-to-one correspondence with the blocks in the image.}
    \label{fig:queries}
\end{figure}

Fig. \ref{fig:vis} visualizes some segmentation results. Fig. \ref{fig:vis} (a) and (d) represent single-target samples, whereas the others are multi-target samples. Fig. \ref{fig:vis} (f) illustrates more challenging sample featuring expressions with multiple category targets and small targets. As depicted in Fig. \ref{fig:vis}, compared with ReLA\cite{liu2023gres}, our approach performs better in capturing object details, understanding the textual information and identifying the corresponding referent. Additionally, our method has more accurate spatial awareness, demonstrating better comprehension of the expressions involving order and orientation. For the multi-category targets in Fig. \ref{fig:vis} (f), our method accurately distinguishes between different classes of targets (person and phone), whereas ReLA confuses the two targets on the right side of the image. Moreover, for small targets in the image, our regional supervision enables the generation of unique class prototypes for each small block, allowing our method to finely delineate the referent's outline.

In Fig. \ref{fig:queries}, we apply the classic unsupervised clustering method, KMeans, to the mask embeddings from the final layer of the model after inputting the image-text pair. These embeddings correspond, one-to-one, with regions, and are used to form a final mask by computing inner products with visual features. Therefore, they can be considered model category prototypes for the corresponding regions in the image. Blocks of the same color indicate that the embeddings for these regions are clustered into the same category, implying that they serve as prototypes for the same category. For clarity, we present two multi-target samples, where Fig. \ref{fig:queries} (a) represents a multi-category sample, and Fig. \ref{fig:queries} (b) represents a single-category sample. In both cases, the categories of our clustered queries exhibit clear regional correspondences. In Fig. \ref{fig:queries} (a), queries for ``cow" and ``person" are handled by two distinct query clusters, whereas in Fig. \ref{fig:queries} (b), the two instances belonging to the same category (cat) are handled by queries from the same cluster. However, in both (a) and (b), the queries decoded by ReLA do not exhibit clear patterns, demonstrating that our regional supervision strategy effectively assigns corresponding class prototypes to different category instances and reduces task complexity.

In both cases, the categories of our clustered queries exhibit clear regional correspondences. In Fig. \ref{fig:queries} (a), queries for the ``cow" and ``person" are handled by two distinct query clusters, whereas in Fig. \ref{fig:queries} (b), the two instances belonging to the same category (cat) are handled by queries from the same cluster. However, in both (a) and (b), the queries decoded by ReLA do not exhibit clear patterns, demonstrating that our regional supervision strategy effectively assigns corresponding class prototypes to different category instances and reduces task complexity.

\section{Limitations}

\begin{figure}[htbp]
    \centering
    \includegraphics[width=0.9\linewidth]{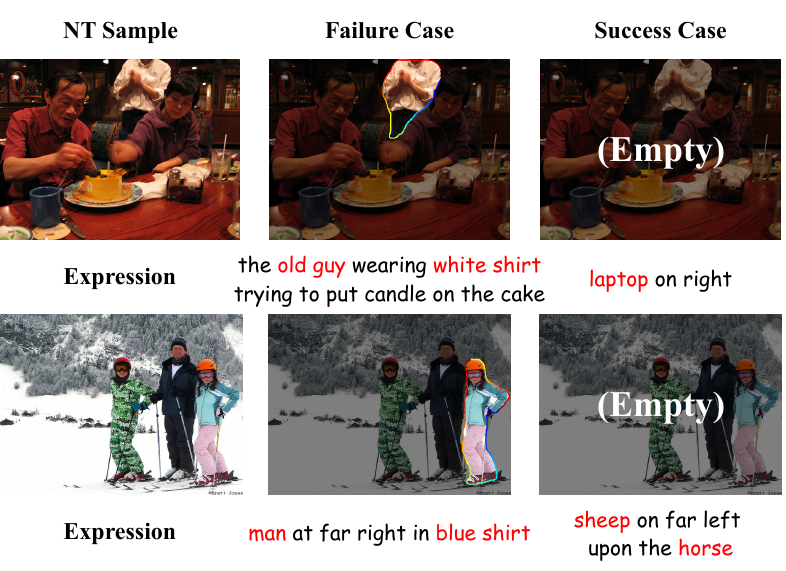}
    \caption{Visualization of several failure cases. Herein, failure cases and their corresponding success cases for two groups of No-Target samples are presented. The first column shows the original images, the failure cases are presented in the second column, and the success cases are presented in the third column. The key nouns in the expressions that determine the target are highlighted in red.}
    \label{fig:limit}
\end{figure}

As shown in Fig. \ref{fig:limit}, the difficulty of  the negative samples in GRES varies greatly. In some easy samples, such as the success cases in Fig. \ref{fig:limit}, the elements mentioned in the expressions, such as ``laptop", ``sheep", and ``horse", are entirely absent from the image, and even contradict the scene. However, in some challenging samples, such as the failure cases in Fig. \ref{fig:limit}, elements such as ``old guy", ``white shirt", and ``man" are present in the image but with slight differences in detail. Dealing with these challenging samples requires consideration of both the understanding capabilities and the long-tail distribution of the NT samples, and it has a substantial effect on the GRES. Therefore, addressing the issue of handling difficult no-target samples will be crucial for solving GRES in future research.

Furthermore, in our model, despite the adaptive binding of prototypes, we still employ a hard-split strategy for features. As shown in the visualization in Fig. \ref{fig:queries}, the regions we construct are relatively coarse-grained with fixed boundaries that cannot be altered, preventing the prototypes from selecting positive samples at the pixel level. Consequently, when the image is excessively complex, or when the region is located at the classification boundary, the challenge persists in dealing with one prototype corresponding to instances of multiple categories. Therefore, it is also worth attempting to implement deformable partitioning of regions or adaptive merging of prototypes.

\section{Conclusion}
In this paper, we reevaluated the distinctions between RES and GRES, and emphasized the heightened difficulty introduced by the scenario in GRES, where multiple instances of different categories are collectively treated as foregrounds. To address this challenge, we proposed a model capable of adaptively binding prototypes. By partitioning the feature map into multiple subregions and supervising them separately, our model dynamically bound prototypes to instances of various categories or different parts of the same instance. Additionally, we designed a mixed modal decoder to better adapt to the no-target samples and extracted class prototypes in GRES. During query initialization, our query generator effectively combined linguistic features to generate region-text-specific initial queries, providing high flexibility. Our proposed model outperformed the current SOTA methods on all three splits of the gRefCOCO dataset and the classical RES datasets RefCOCO+ and G-Ref. It also achieved very competitive results on the classical RES dataset RefCOCO.

\newpage
\bibliographystyle{IEEEtran}
\bibliography{name.bib}

\vspace{-33pt}
\begin{IEEEbiography}[{\includegraphics[width=1in,height=1.25in,clip,keepaspectratio]{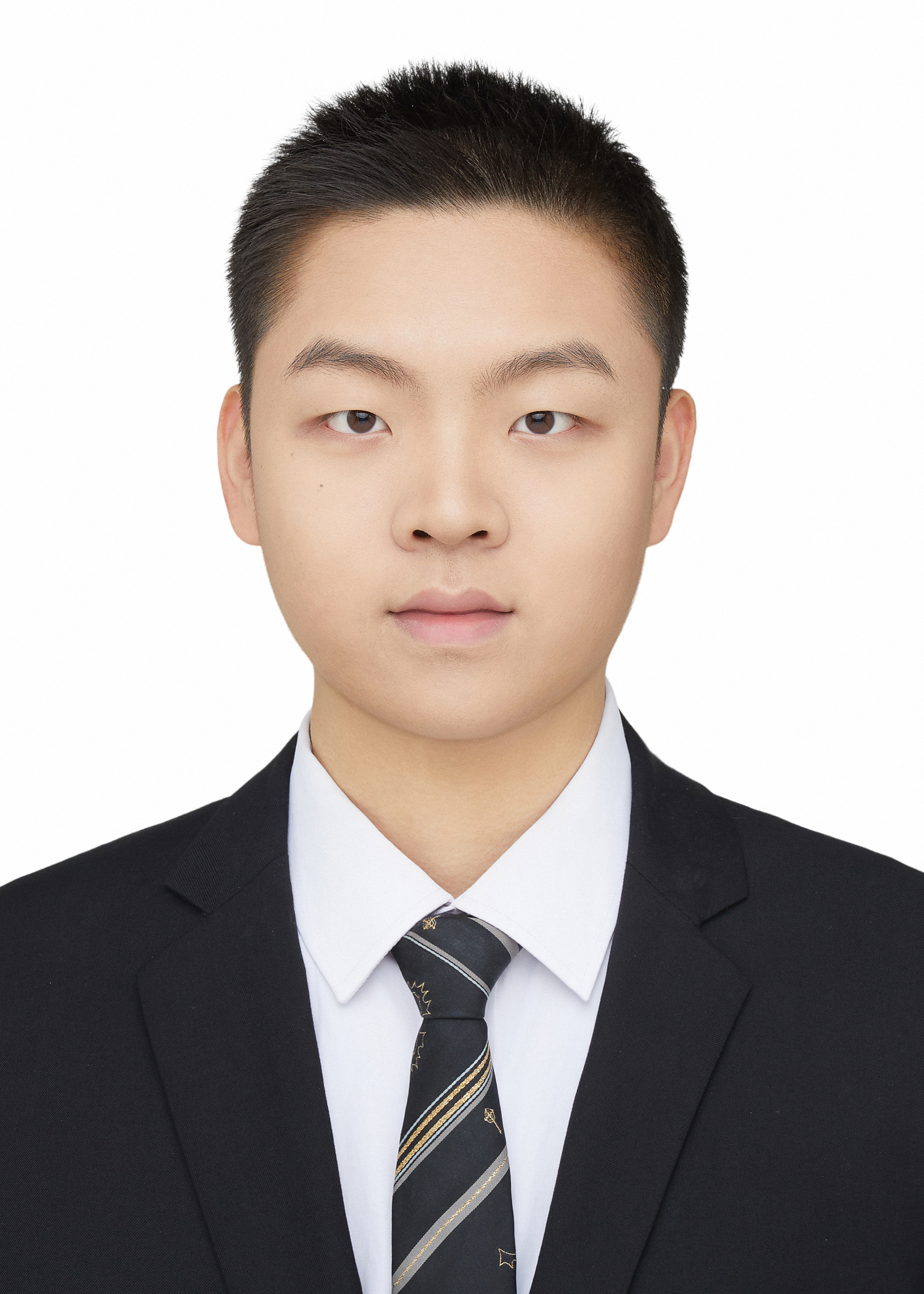}}]{Weize Li}
is currently pursuing the Ph.D. degree with the Beijing Key Laboratory of Network System and Network Culture, Beijing University of Posts and Telecommunications, Beijing, China. His current research interests include computer vision, object perception and cross-modal Learning.
\end{IEEEbiography}

\begin{IEEEbiography}
[{\includegraphics[width=1in,height=1.25in,clip,keepaspectratio]{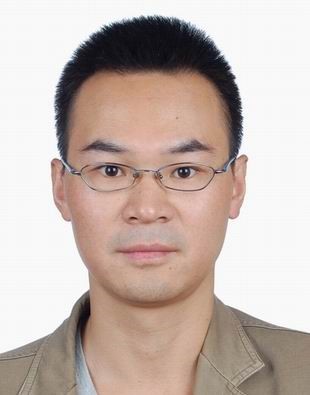}}]%
{Zhicheng Zhao} received the Ph.D. degree in communication and information systems from the Beijing University of Posts and Telecommunications, China, in 2008. He was a Visiting Scholar with the School of Computer Science, Carnegie Mellon University, USA, from 2015 to 2016. He is currently a Professor with the Beijing University of Posts and Telecommunications. He has authored or coauthored more than 150 journal articles and conference papers. His research interests are computer vision, and image and video semantic understanding and retrieval. 
\end{IEEEbiography}

\begin{IEEEbiography}[{\includegraphics[width=1in,height=1.25in,clip,keepaspectratio]{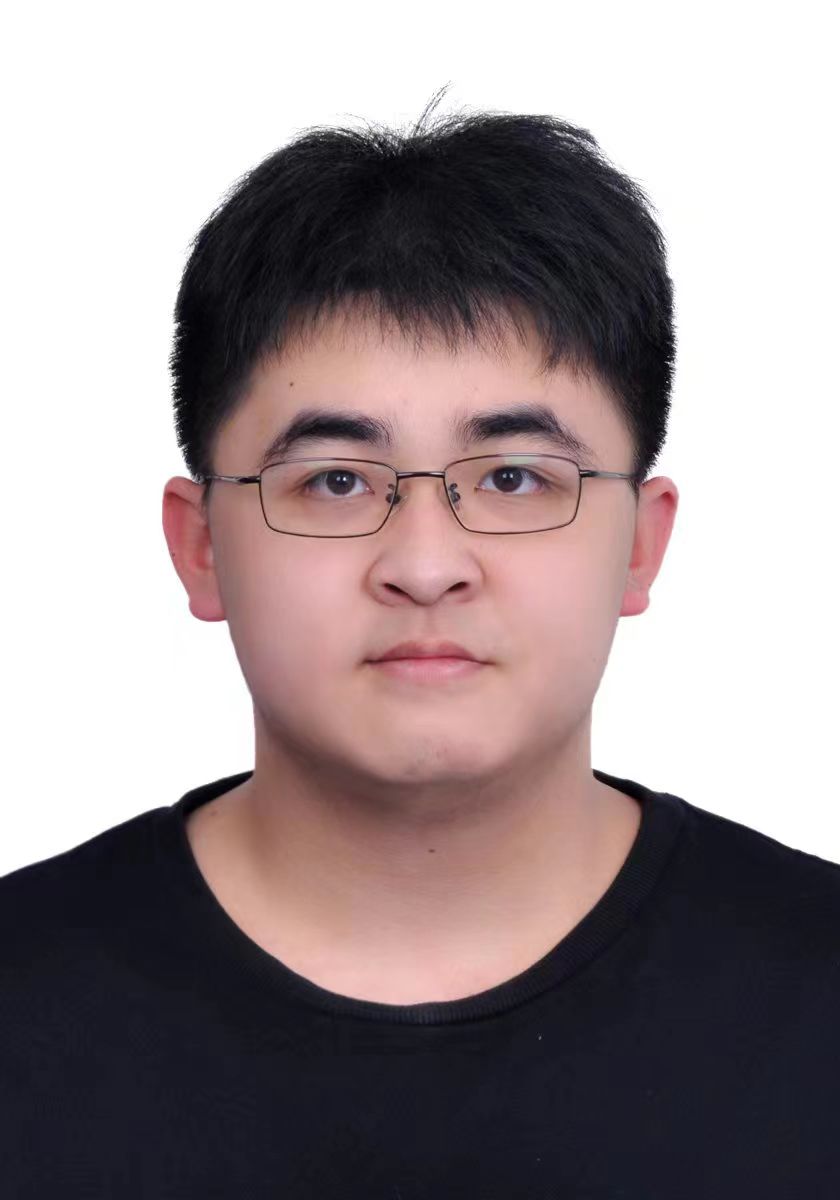}}]{Haochen Bai}
is currently pursuing the M.S degree with the Beijing Key Laboratory of Network System and Network Culture, Beijing University of Posts and Telecommunications, Beijing, China. His recently research direction is computer vision and referring image segmentation.
\end{IEEEbiography}

\begin{IEEEbiography}[{\includegraphics[width=1in,height=1.25in,clip,keepaspectratio]{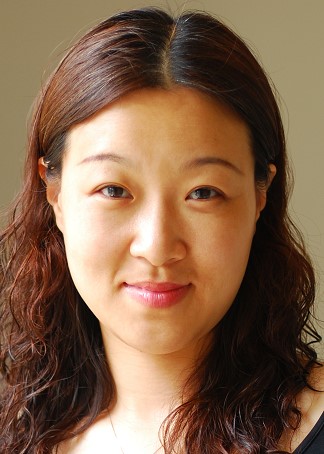}}]%
{Fei Su} received the Ph.D. degree in communication and electrical systems from the Beijing University of Posts and Telecommunications (BUPT), China, in 2000. She was a Visiting Scholar with the Department of Electrical and Computer Engineering, Carnegie Mellon University, USA, from 2008 to 2009. She is currently a Professor with the Multimedia Communication and Pattern Recognition Laboratory, BUPT. She has authored and coauthored more than 150 journal articles and conference papers and some textbooks. Her current interests include pattern recognition, image and video processing, and biometrics.
\end{IEEEbiography}

\vfill

\end{document}